\documentclass[letterpaper]{article} % DO NOT CHANGE THIS
\usepackage{aaai24}  % DO NOT CHANGE THIS
\usepackage{times}  % DO NOT CHANGE THIS
\usepackage{helvet}  % DO NOT CHANGE THIS
\usepackage{courier}  % DO NOT CHANGE THIS
\usepackage[hyphens]{url}  % DO NOT CHANGE THIS
\usepackage{graphicx} % DO NOT CHANGE THIS
\usepackage{natbib}  % DO NOT CHANGE THIS AND DO NOT ADD ANY OPTIONS TO IT
\usepackage{caption} % DO NOT CHANGE THIS AND DO NOT ADD ANY OPTIONS TO IT
\DeclareCaptionStyle{ruled}{labelfont=normalfont,labelsep=colon,strut=off} % DO NOT CHANGE THIS
\usepackage{amsfonts}
\usepackage{amsmath}
\usepackage{amssymb}
\usepackage{amsthm}
\usepackage{mathtools}
\usepackage{stmaryrd}

\usepackage{cleveref}
\usepackage{xcolor}

\usepackage{algorithm}
\usepackage[noend]{algpseudocode}
\usepackage{subcaption}

\newcommand{\hide}[1]{}

\DeclareMathOperator*{\argmax}{arg\,max}

\newtheorem{prop}{Proposition}

\usepackage{newfloat}
\usepackage{listings}
\floatstyle{ruled}
\newfloat{listing}{tb}{lst}{}
\floatname{listing}{Listing}
\title{Learning to Learn in Interactive Constraint Acquisition}
\author{
    Dimos Tsouros\textsuperscript{\rm 1}, 
    Senne Berden\textsuperscript{\rm 1}, 
    Tias Guns\textsuperscript{\rm 1}%\thanks{With help from the AAAI Publications Committee.}\\
}
\affiliations{
    \textsuperscript{\rm 1}KU Leuven, Belgium\\
    dimos.tsouros@kuleuven.be, senne.berden@kuleuven.be, tias.guns@kuleuven.be
}

\begin{document}

\maketitle

\begin{abstract}
Constraint Programming (CP) has been successfully used to model and solve complex combinatorial problems. However, modeling is often not trivial and requires expertise, which is a bottleneck to wider adoption. %As a result, the field of Constraint Acquisition (CA) has evolved with the aim to (semi-)automate the modeling process by combining CP and %(symbolic) 
%Machine Learning (ML). 
%
%OLD:In Constraint Acquisition (CA), the system combines CP and Machine Learning (ML) to assist the user in the modeling process.
%In (inter)active CA, the system interacts with the user while learning the model, by posting queries that the user has to answer.
In Constraint Acquisition (CA), the goal is to assist the user by automatically \textit{learning} the model.
In (inter)active CA, this is done by interactively posting queries to the user, e.g., asking whether a partial solution satisfies their (unspecified) constraints or not.
While interactive CA methods learn the constraints, the learning is related to symbolic concept learning, as the goal is to learn an exact representation. 
However, a large number of queries is still required to learn the model, %to converge to the set of constraints modeling the problem the user has in mind,
which is a major limitation.
%Interactive CA uses search-based learning which is often not able to detect relevant patterns.
In this paper, we aim to alleviate this limitation by tightening the connection of CA and Machine Learning (ML), by, for the first time in interactive CA, exploiting statistical ML methods. %by proposing  to exploit statistical ML techniques to guide the acquisition process, %in the most promising parts of the problem, 
%aiming to ask queries that can give more information to the system, leading to a lower amount of queries to converge.
We propose to use probabilistic classification models to guide interactive CA to generate more promising queries. We discuss how to train classifiers to predict whether a candidate expression from the bias is a constraint of the problem or not, using both relation-based and scope-based features. We then show how the predictions can be used in all layers of interactive CA: 
the query generation, the scope finding, and the lowest-level constraint finding.
%TIAS THINKS THE ABSTRACT IS TOO LONG we first show how to integrate them in the objective function of the state-of-the-art query generator \textsc{PQ-Gen}. We then propose a modification in the function for finding the scope of the constraints, to exploit the predictions from the previous step. Finally, we show how the (modified) objective function from the first step can be used in the last layer of CA, namely \textsc{FindC}.
%% to guide the subsequent queries. %Finally, we show how the (modified) objective function from the first step can be exploited in the last layer of CA, namely \textsc{FindC}.
We experimentally evaluate our proposed methods using different classifiers and show that our methods greatly outperform the state of the art, decreasing the number of queries needed to converge by up to 72\%.

%with The goal is to minimize the number of interactions, while the waiting time for the user remains reasonable.

\end{abstract}

%and also evaluate the performance of CA when the predicted class is used directly instead of the probabilities.

\section{Introduction}

Constraint Programming (CP) is considered one of the foremost paradigms for solving combinatorial problems in Artificial Intelligence. In CP, the user declaratively states the constraints over a set of decision variables, defining the feasible solutions to their problem, and then a solver is used to solve it. 
%The basic assumption in constraint programming (CP) is that the user first models the problem and a solver is then used to solve it. 
Although CP has many successful applications on combinatorial problems from various domains, the modeling process is not always trivial and is limiting non-experts from using CP on complex problems. This is considered a major bottleneck for the wider adoption of CP~\cite{freuder2014grand,freuder2018progress}. %\red{Expressing a combinatorial problem as a set of constraints over decision variables requires substantial expertise}~\cite{freuder1999modeling}.

Motivated by the need to overcome this obstacle, assisting the user in \mbox{modeling} is regarded as an important research topic~\cite{kolb2016learning,de2018learning,freuder2018progress,lombardi2018boosting}. In {\em Constraint Acquisition (CA)}, which is an area where CP meets Machine Learning (ML), the model of a constraint problem is learned from a set of examples (i.e., assignments to the variables) of solutions, and possibly non-solutions. %The set of examples is either pre-existing and given to the system by the user, or generated by the system and posted to the user to label them.
%State-of-the-art CA systems use the candidate elimination and version space paradigms~\cite{mitchell1978version,mitchell1997concept}.
%CA can come in various flavors, depending on factors such as whether the learner can post queries to the user dynamically, and the type of queries that can be posted and answered.

In {\em passive} CA, a set of pre-existing examples is given to the system, and using these examples a set of constraints is returned~\cite{bessiere2004leveraging,bessiere2005sat,lallouet2010learning,beldiceanu2012model,bessiere2017constraint,kumar2022learning,berden2022learning}. %Many different approaches have been devised. 
%\textsc{Conacq.1} is a version space algorithm for learning fixed-arity constraints~\cite{bessiere2005sat,bessiere2017constraint}, while ModelSeeker learns global constraints% that are taken from a predefined constraint catalog
%~\cite{beldiceanu2012model}. Recently, COUNT-CP was introduced, which is a generate-and-aggregate approach that can learn expressive first-order constraints \cite{kumar2022learning}. Aiming to deal with noise in the data, in~\cite{prestwich2020robust} a statistical approach based on sequential analysis named \textsc{SeqAcq} was proposed, while in~\cite{prestwich2021classifier}, a naive Bayes classifier is trained over the set of solutions and non-solutions of the problem, from which a constraint network is derived.

On the other hand, {\em active} or {\em interactive} acquisition systems interact with the user to learn a target set of constraints, which represent the problem the user has in mind~\cite{freuder1998suggestion,bessiere2007query,bessiere2017constraint}. In the early days, most methods only made use of membership queries (is this a solution or not?)~\cite{angluin1988queries,bessiere2007query}, while a more recent family of algorithms also makes use of \textit{partial membership queries}~\cite{arcangioli2016multiple,bessiere2013constraint,lazaar2021parallel,tsouros2020efficient,tsouros2021learning,tsouros2019structure,tsouros2020omissions,mquacq}. Such (partial) queries ask the user to classify (partial) assignments to the variables as (non-)solution. %A recent advancement in active CA is a meta-algorithm named \textsc{GrowAcq}~\cite{tsouros2023guided}, which is able to use any other algorithm to learn the constraints in a bottom-up fashion. This allowed systems to handle significantly larger sets of candidate constraints and reduced the maximum waiting time for the user.\senne{Aren't we favoring our own work a bit too much in the previous sentence? \textsc{GrowAcq} isn't specifically relevant to this paper, so I wouldn't talk about it in this introduction too much}\dt{GrowAcq makes it possible to guide query generation, without growacq no time is left to actually optimize (any) objective function. But we could remove it from intro (or just mention) and say more in background etc} 
Recently, a way to guide the top-level query generation was introduced~\cite{tsouros2023guided}, based on counting-based probabilistic estimates of whether candidate expressions are constraints of the problem or not. Using this method, the number of queries required to converge decreased significantly. 

Despite the recent advancements in active CA, there are still significant drawbacks to overcome. One of the most important drawbacks is the large number of queries still required in order to find all constraints. We believe this is due to the search-based learning being mostly \textit{uninformed}. During learning it is not aware of patterns that may appear in the constraints acquired so far, which can guide the rest of the process. An exception is the \textsc{Analayze\&Learn}~\cite{tsouros2019structure} function, which tries to detect potential cliques in the constraint network learned.

In this work, we focus on this major limitation and contribute the following elements to alleviate it:

\begin{itemize}

    \item We show how probabilistic classification can be used to predict whether a candidate expression is a constraint of the problem or not, based on the constraints learned so far and the ones removed from the candidate set at any point during the acquisition process. We use both relation-based and scope-based features to train ML models that are then exploited to guide interactive CA systems.
    
    \item Previously it was shown that top-level query generation can be guided with (counting-based) probabilistic estimates. We show how such guidance can be extended to all layers of interactive CA where queries are asked. %We propose a small modification in \textsc{FindScope} function, converting the problem of (randomly) splitting the variables in half to a CP problem incorporating an objective function to select the most promising next query. We also show how to use the predicted probabilities to guide the \textsc{FindC} function, which is used to learn the specific relation(s) in the scope found.  
    
    %\item We discuss how the predictions from (probabilistic) classification systems can be used when guiding interactive CA.
     
%    \item We show how the predictions can be exploited to guide all the layers of interactive CA systems, integrating them into the objective function of the state-of-the-art query generator \textsc{PQ-Gen}, in our proposed \textsc{G-FindScope} function and in the \textsc{FindC} function that is used to learn the specific relation(s) in the scope found.    
    
    \item We make a comprehensive experimental evaluation of our proposed methods, showing the effect of different classifiers, focusing on the number of queries vs. runtime for the ML-guided systems. We also show the effect of guiding all layers where queries are posted to the user. %We also show that classification metrics are not always aligned with the improvement each classifier offers in the acquisition process.
\end{itemize}

%The rest of the paper is structured as follows. First, some background on CA is given. ...An experimental evaluation of the methods is given in Section~\ref{sec:exp}, and is followed by conclusions~\Cref{sec:concl}. %Finally, Section~\ref{sec:concl} concludes the paper.

\section{Background}
\label{sec:back}

%In this section, we give the necessary background on constraint satisfaction problems and CA.
Let us first give some basic notions regarding constraint satisfaction problems.

A \textit{constraint satisfaction problem} (\textit{CSP}) is a triple $P = (X, D, C)$, consisting of:

\begin{itemize}

\item a set of $n$ variables $X = \{x_1, x_2, ..., x_n\}$, representing the entities of the problem,

\item a set of $n$ domains $D = \{D_1, D_2, ..., D_n\}$, where $D_i \subset \mathbb{Z}$ is the finite set of values for $x_i$,

\item a constraint set (also called constraint network) $C = \{c_1, c_2, ..., c_t\}$.

\end{itemize}

A \textit{constraint} $c$ is a pair ($rel(c)$, $var(c)$), where $var(c)$ $\subseteq X$ is the \textit{scope} of the constraint, and $rel(c)$ is a relation over the domains of the variables in $var(c)$, that (implicitly) specifies which of their value assignments are allowed. $|var(c)|$ is called the \textit{arity} of the constraint. The constraint set $C[Y]$, where $Y \subseteq X$, denotes the set of constraints from $C$ whose scope is a subset of $Y$. The set of solutions of a constraint set $C$ is denoted by $sol(C)$. 
%A \textit{redundant} or \textit{implied} constraint $c$ in $C$ is a constraint such that $sol(C) = sol(C\setminus \{c\})$. 

An \textit{example} $e_Y$ is an assignment on a set of variables $Y \subseteq X$. $e_Y$ is rejected by a constraint $c$ iff $var(c)$ $\subseteq Y$ and the projection $e_{var(c)}$ of $e_Y$ on the variables in the scope $var(c)$ of the constraint is not in $rel(c)$.
A complete assignment $e$ that is accepted by all the constraints in $C$ is a \textit{solution} to $C$, i.e., $e \in sol(C)$. An assignment $e_Y$ is called a \textit{partial solution} iff $e_Y \in sol(C[Y])$. %Note that a partial solution is not necessarily part of a complete solution. 
$\kappa_C(e_Y)$ represents the subset of constraints from a constraint set $C[Y]$ that reject $e_Y$.
%, while $\lambda_C(e_Y)$ denotes the set of constraints in $C[Y]$ that satisfy $e_Y$.

In CA, the pair $(X, D)$ is called the \textit{vocabulary} of the problem at hand and is common knowledge shared by the user and the system. Besides the vocabulary, the learner is also given a \textit{language} $\Gamma$ consisting of {\em fixed arity} constraints. Using the vocabulary $(X, D)$ and the constraint language $\Gamma$, the system generates the \textit{constraint bias} $B$, which is the set of all expressions that are candidate constraints for the problem. The (unknown) target constraint set $C_T$ is a constraint set such that for every example $e$ it holds that $e \in sol(C_T)$ iff 
$e$ is a solution to the problem the user has in mind.
%$ASK(e) = True$. %Even though the user does not know $C_T$, we assume that there exists a constraint set that captures the model the user has in mind but cannot express in a mathematical formulation. 
The goal of CA is to learn a constraint set $C_L$ that is equivalent to the target constraint set $C_T$.

%The first condition captures that $\kappa_B(e_Y)$ cannot be empty. If $\kappa_B(e_Y)$ would be empty, then the answer to the query $ASK(e_Y)$ could not be negative, based on the assumption that $C_T$ is representable by $B$. The second condition captures the fact that $e_Y$ should not be rejected by any constraint in the learned network $C_L$, since otherwise it would certainly be classified as negative. Following the literature, we assume that all queries are answered correctly by the user.

\subsection{Interactive Constraint Acquisition}

In interactive CA, the system interacts with the user while learning the constraints. The classification question $ASK(e_X)$, asking the user if a complete assignment $e_X$ is a solution to the problem that the user has in mind, is called a \textit{membership query}~\cite{angluin1988queries}. 
%The answer of the user to a membership query is positive if $e_X \in sol(C_T)$ and negative otherwise. 
A \textit{partial query} $ASK(e_Y)$, with $Y \subset X$, 
asks the user to determine if $e_Y$, which is an assignment in $D^Y$, 
is a \textit{partial solution} or not, i.e., if $e_Y \in sol(C_T[Y])$. 
A (complete or partial) query $ASK(e_Y)$ is called \textit{irredundant} iff the answer is not implied by information already available. % to the system. 
That is, it is irredundant iff $e_Y$ is rejected by at least one constraint from the bias $B$, and not rejected by the network $C_L$ learned thus far. 

%\begin{equation}
%\label{eq:qgen}
%\begin{aligned}
%Ask(e_Y) \text{ is irredundant} \iff e_Y \in sol( C_L[Y] \wedge \bigvee_{c_i \in B[Y]} \neg c_i ),
%\end{aligned}
%\end{equation}
%CA systems should ask only irredundant queries.

% Generic pseudocode -------------------------------------------------
\begin{algorithm}[t]
\caption{Generic Constraint Acquisition Template}\label{alg:general}
\begin{algorithmic}[1]

\Require $X$, $D$, $B$, $C_{in}$ ($X$: the set of variables, $D$: the set of domains, $B$: the bias, $C_{in}$: an optional set of known constraints)
\Ensure $C_L$ : the learned constraint network

\State $C_L \leftarrow C_{in}$

\While {True}

	\State $e \leftarrow$ \textsc{QGen}($C_L$, $B$) %Generate an $e$ accepted by $C_L$ and rejected by $B$

	\If{ $e$ = nil } \Return $C_L$  \Comment{converged}
	\EndIf
    
    \If{$ASK(e)$ = True}     %\Comment{Post $e$ to the user to classify it}
        \State $B \leftarrow B \setminus \kappa_B(e)$ %Remove constraints rejecting $e$ from $B$
    \Else
        \State $(B,S) \leftarrow$ \textsc{FindScope}($e$, $B$) %Find the scope $var(c)$ of a $c \in \kappa_{C_T}(e)$
        \State $(B,C_L) \leftarrow$ \textsc{FindC}($S$, $C_L$, $B$)%Find all $c' \in C_T \mid var(c) = var(c')$
    \EndIf

\EndWhile

\end{algorithmic}
\end{algorithm}

% End of Generic pseudocode ------------------------------------------

\Cref{alg:general} presents the generic process followed in interactive CA through partial queries. The learned set $C_L$ is first initialized either to the empty set or to a set of constraints given by the user that is known to be true (line 1). Then the main loop of the acquisition process begins, where first the system generates an irredundant example (line 3) and posts it as a query to the user (line 5). If the query is classified as positive, then the candidate expressions from $B$ that violate it are removed (line 6). 
If the example is classified as negative, then the system tries to find one (or more) constraint(s) from $C_T$ that violates it. This is done in two steps. First, the scope of one or more violated constraints is found, by asking queries and possibly shrinking the bias along the way (line 8). Then, the relations of the constraints in this scope(s) are found, again by asking queries and possibly shrinking the bias (line 9). %Both of these steps involve posting additional partial queries to the user. 

This process continues until the system converges. The acquisition process has \textit{converged} on the learned network $C_L \subseteq B$ iff $C_L$ agrees with the set of all labeled examples $E$, and for every other network $C \subseteq B$ that agrees with $E$, it holds that $sol(C) = sol(C_L)$. This is proved if no example could be generated at line 3, as in this case, all constraints in $B$ are redundant.

Notice that, interactive CA systems consist of three components where (increasingly simpler) queries are asked to the user: (1) Top-level query generation (line 3), (2) Finding the scope(s) of violated constraints (line 8), (3) Finding the relations of constraints in the scopes found (line 9).

%If the first condition is true ($C_L$ agrees with $E$) but the second condition has not been proved when the acquisition process finishes, then we have \textit{premature convergence}. This can happen when we return at line 3 without an example, without proving that an irredundant example does not exist (e.g., because of a time limit).

State-of-the-art algorithms like QuAcq~\cite{bessiere2013constraint,bessiere2023learning}, MQuAcq~\cite{tsouros2020efficient} and MQuAcq-2~\cite{tsouros2019structure} follow this template. %The functions commonly used to locate the scope of a constraint (line 8) are \textit{FindScope}~\cite{bessiere2013constraint} or the more efficient \textit{FindScope-2}~\cite{tsouros2020efficient}. In these functions, subsets of variables are recursively removed and the resulting partial example is posted as a query again. If the answer of the user changes to ``yes'', then the system knows that at least one of the removed variables belongs to the scope of a violated constraint. This process continues until the full scope is found. However, no guidance is used on what subset of variables to remove or to keep in each subsequent query. \dt{Move the explanation to the section for guiding}
%To learn the constraints in the scope found (line 9), the \textit{FindC} function is used~\cite{bessiere2013constraint,bessiere2023learning}. 
Recently, a \textit{meta-}algorithm named \textit{GrowAcq}~\cite{tsouros2023guided} %(Algorithm~\ref{alg:growacq})
was introduced, in order to handle large biases and to reduce the number of queries. The key idea is to call a CA algorithm on an increasingly large subset of the variables $Y \subseteq X$, initially using a small number of variables, each time using a (growing) subset of the potentially huge bias. %Using this method, CA systems are able to handle large sets of candidate constraints, simplifying the query generation process to take into account smaller sets of constraints. %This makes query generation faster and thus able to use more advanced methods in the objective function without delaying presenting a query to the user.

\subsection{Guiding Query Generation}
When using \textsc{GrowAcq}, only a subset of $B$ needs to be considered at a time, and query generation is often fast, leaving sufficient room for using optimization to find a \textit{good} query in top-level query generation (line 3 of~\Cref{alg:general}). Query generation is formulated as a CSP with variables $Y$ and constraints $ C_L[Y] \wedge \bigvee_{c_i \in B[Y]} \neg c_i$, in order to find an example $e_Y$. % satisfying the properties of~\Cref{eq:qgen}.
Hence, when the set of candidates $B$ is reduced, query generation is simplified.

As a result of this speed-up, in~\cite{tsouros2023guided} a method to guide the top-level query generation was proposed. This method introduces an objective function that uses the prediction of a model ${\cal M}(c)$:

\begin{equation}
\label{eq:guideqgen}
\begin{aligned}
e = \argmax_e 
\sum_{c \in B} \llbracket e \not\in sol(\{c\}) \rrbracket \cdot (1 - |\Gamma| \cdot \llbracket {\cal M}(c) \rrbracket)
\end{aligned}
\end{equation}
where $ \llbracket \cdot \rrbracket$ is the Iverson bracket which converts \textit{True}/\textit{False} into 1/0%(this notation will be reused later in the paper)
.

The objective function's aims are twofold. First, it wants queries that lead to a positive answer to violate many constraints in the bias, shrinking it faster. Second, it wants constraints that lead to a negative answer to violate a small number of constraints from the bias, so that the actual constraint leading to the negative answer can be found more easily. For more exposition on how this objective function achieves these aims, we refer the reader to~\cite{tsouros2023guided}.
% The goal of this objective function is to minimize the violation of constraints that are in the unknown target set $C_T$ (by giving a ``penalty'' when violated of size $|\Gamma|$, which is equal to the upper bound of the number of constraints in each scope), seeking a query to which the user's answer will be `yes'. In addition, the goal is also to maximize the violation of constraints in $B$ that are not in $C_T$ (by increasing the value of the objective function by 1 when violated). Thus, positive answers can shrink the bias faster, while in negative queries a minimum number of constraints will be violated, making it easier to find out which one of them is the constraint we seek. 

% On the one hand, every time that the oracle returns \textit{False} for a constraint from the bias that is violated by $e$, the objective function is increased by 1, thereby maximizing the violation of these constraints.
% Conversely, for constraints where ${\cal O}$ returns \textit{True}, we aim to minimize the violations, which requires a reduction in the objective value for each such violated constraint, giving a ''penalty'' of size $|\Gamma|$, which is equal to the upper bound of the number of constraints in each scope. 
% This ensures that the system prioritizes satisfying a constraint with  ${\cal O}(c_j) =$ \textit{True}, over violating other constraints from $B$.

Model ${\cal M}$ tries to determine for every constraint $c$ whether violating or satisfying $c$ would lead to the least amount of queries later on in the algorithm, based on a probabilistic estimate $P(c \in C_T)$ of how likely a constraint is to belong to the target set of constraints of the problem

\begin{equation}
\label{eq:oracle}
\begin{aligned}
{\cal M}(c) = \big( \frac{1}{P(c \in C_T)} \leq log(|Y|) \big)
\end{aligned}
\end{equation}

\section{Using Probabilistic Classification to Guide Interactive CA}
%story
% IN [] a method to guide top-level query generation in interactive constraints was proposed. In this method .....
% Oracle basically classifies constraints as true fals
% use machine learning techniques instead of simple counting
% Subsection: Dataset
% During constraint acquisition we learn some constraints and we discard some others -> we can use information from these to predict for the remaining ones
% what information can we use? features of constraints (table)....

%We have discussed how we can use a pseudo-oracle that predicts if a constraint is part of the problem in order to guide all layers of interactive constraint acquisition. 
%The pseudo-oracle utilized to guide query generation is using a probabilistic estimate of how likely a candidate constraint is to be part of the problem. 

The model ${\cal M}$ leverages a probabilistic estimate of the likelihood of a given candidate constraint belonging to the problem at hand.
In~\cite{tsouros2023guided}, a simple counting-based method was utilized that only uses information about the relation of the constraints. That is, the number of times a constraint with relation $rel(c)$ has been added to $C_L$ is counted, and then divided by the total number of times that such a constraint has been removed from $B$.

While this technique provides basic guidance, we propose to use more advanced prediction techniques. Specifically, we propose to use statistical machine learning techniques, exploiting probabilistic classification in order to calculate $P(c \in C_T)$.

%While the technique detailed above provides basic guidance, we propose to use more advanced probability estimation techniques. Specifically, we propose to use statistical machine learning techniques, exploiting probabilistic classification in order to calculate $P(c \in C_T)$.

In order to use probabilistic classification in this context, we need to build a dataset to learn from. We formally define a dataset $\mathcal{D}$ as a collection of $N$ instances, each instance corresponding to a constraint. Each instance is a tuple $(\mathbf{x}_i, y_i), i \in {1, 2, ..., N}$,  with $\mathbf{x}_i$ being a vector of features of constraint $c_i$, and $y_i$ being a (Boolean) label that denotes whether $c_i \in C_T$.

\begin{table}[h]
    \centering
    \begin{tabular}{|p{0.015\textwidth}|p{0.125\textwidth}|p{0.042\textwidth}|p{0.22\textwidth}|}
        \hline
        ID & Name & Type & Description \\
        \hline
        1 & Relation & String & Constraint relation \\
        \hline
        2 & Arity & Int & Constraint arity \\
        \hline
        3 & Has\_constant & Bool & If a constant value is present \\
        \hline
        4 & Constant & Int & The constant value \\
        \hline
        5 & Var\_name\_same & Bool & If all variables share the same name \\
        \hline
        6 & Var\_Ndims\_same & Bool & If the number of dimensions of all variables is the same \\
        \hline
        7 & Var\_Ndims\_max & Int & The maximum number of dimensions among variables \\
        \hline
        8 & Var\_Ndims\_min & Int & The minimum number of dimensions among variables \\
        \hline
        9 & Var\_dim$_i$\_has & Bool & If dimension $i$ is present for all variables\\
        \hline
        10 & Var\_dim$_i$\_same & Bool & If a specific dimension of all variables is the same \\
        \hline
        11 & Var\_dim$_i$\_max & Int & Maximum dimension $i$ value among variables \\
        \hline
        12 & Var\_dim$_i$\_min & Int & Minimum dimension $i$ value among variables \\
        \hline
        13 & Var\_dim$_i$\_avg & Float & Average dimension $i$ value among variables \\
        \hline
        14 & Var\_dim$_i$\_spread & Float & Spread of dimension $i$ values among variables \\
        \hline
    \end{tabular}
    \caption{Features for each constraint}
    \label{tab:constraint-features}
\end{table}

To be able to use constraints as instances in our dataset, we need to have a feature representation of constraints. In this paper, for the feature representation, we use both relation-based and scope-based information, exploiting the information we have for the constraint's relation, the variables of its scope, their indices, name, etc. The features we use are shown in~\Cref{tab:constraint-features}.
Note that variables can be given to the CA system in the form of a matrix or tensor. For example, a natural way to structure the variables representing the cell assignments Sudoku is in a 9x9 2-dimensional matrix. When variables are given in such a form, we represent in the features information about the indices of the variables occurring in the constraints, in each dimension of the tensor they were given in. This allows the system to detect patterns like all variables occurring in the same row or column, not being spread out in some dimension, etc., which are common patterns in CP problems.

The dataset $\mathcal{D}$ is grown incrementally throughout the CA process, as gradually more information is obtained about constraints from the initial bias. More concretely, whenever a constraint is removed from $B$ (because they were verified to not be part of $C_T$), it is added to $D$ with a negative label. On the other hand, whenever a constraint gets added to $C_L$, it also gets added to $D$ with a positive label.
%This information can be leveraged to train ML models, which in turn facilitate the prediction of the remaining candidates residing in $B$ via binary classification.
%We can use this information to train ML models so we can derive predictions for the remaining candidates in $B$ using binary classification. 

%The feature vector $\mathbf{x}_i$ represents the characteristics or attributes associated with each particular instance, while $y_i$ denotes the class membership of the instance.

%In the context of constraint acquisition, where we want to predict the probability of a constraint being part of the target network of the problem, each instance represents a constraint that we want to classify as True or False. An instance has a positive label if it was added to $C_L$ and a negative label if it was removed from $B$.

%Probabilistic classifiers like Naive Bayes, or Neural Networks, 
%can make probabilistic estimates after training on binary-labeled data. These probabilities will be used for $P(c \in C_T)$ in \Cref{eq:guideqgen}.

Since dataset $\mathcal{D}$ grows throughout the CA process, the probabilistic classifier should be updated regularly. How often this update should be performed is an important question, as this may affect the waiting time for the user when interacting with the CA system. In this paper, we retrain the classifier on the current dataset $\mathcal{D}$ right before every top-level query generation (line 3 of \Cref{alg:general}), exploiting all the collected information each time. Preliminary experiments showed that this yields the best results.

%\dt{previously discussed stories}
%\dt{story1: discuss query generation objective function - oracle proposed in IJCAI paper - more advanced probabilities estimator - why probabilistic classification fits in this - counts method can be considered a simple probabilistic classification method (with 1 feature) - use machine learning instead of counts - discuss features to use - subsection on how to use it on finding the scope of the constraint (FindScope) - subsection on how to find the relation of the constraint (FindC)}

%\dt{story2: discuss guiding CA in general (MQuAcq-2 + IJCAI sumbission), emphasizing on guiding with probabilities (based on Ijcai submission) - more advanced probabilities estimator - why probabilistic classification fits in this - counts method can be considered a simple probabilistic classification method (with 1 feature) - use machine learning instead of counts - discuss features to use - subsection on how in Query Generation - subsection on how to use it on finding the scope of the constraint (FindScope) - subsection on how to find the relation of the constraint (FindC)}

%\dt{Story3: 3 layers of CA, we present new findscope easier to guide, query generation in each. how to guide? same function - final remaining question: how do the probabilities come? we used counts, and now we are proposing the use of ML. - how creating the dataset (every time ... negative/positive) - Features - when to train? Discuss alternatives. before qgen, every X qgen, every GrowAcq iteration.}

\section{Guiding all layers of Interactive CA} 

As~\cite{tsouros2023guided} showed, guiding top-level query generation can reduce the number of queries significantly and improve CA systems. However, CA systems ask queries to the user also in the \textsc{FindScope} (line 8 of ~\Cref{alg:general}) and \textsc{FindC} (line 9 of ~\Cref{alg:general}) components, which respectively try to find the scope of one or more violated constraints, and then all constraints on that scope. %A significant portion of the queries posted comes from these components. 
While guiding the generation of top-level queries delivers significant advantages, neglecting guidance within these two layers is a missed opportunity.

In the rest of this section, we show how to use the same logic for guiding query generation in the remaining two layers of CA systems.

\subsection{Guiding FindScope}

The functions used in the literature~\cite{bessiere2013constraint,tsouros2020efficient,bessiere2023learning} to find the scope of violated constraints after a negative answer from the user (line 8 of~\Cref{alg:general}) work in a similar way.  We will use \textsc{FindScope} from~\cite{bessiere2023learning} (shown in~\Cref{alg:findscope}) to demonstrate our method, but the same logic applies to all existing in the literature. \textsc{FindScope} methods recursively map the problem of finding a constraint to a simpler problem by removing blocks of variable assignments from the original query (the one asked in line 3 of \Cref{alg:general}, to which the user answered ``no'') and asking partial queries to the user. The removed block must contain at least one variable while not including all the present variables, in order to lead to an irredundant query. If after the removal of some variables, the answer of the user changes to ``yes'', then the removed block contains at least one variable from the scope of a violated constraint. When this happens, \textsc{FindScope} focuses on refining this block, adding some variable assignments back to the query. When, after repeating this procedure, the size of the considered block becomes 1 (i.e., the block contains a single variable), this variable is found to be in the scope of a violated constraint we seek (line 5 of \Cref{alg:findscope}). %The main differences between existing functions are regarding their mechanisms to avoid posting redundant queries to the user, and having some recursive calls that do not post a query. 

% Algorithm findScope ------------------------------------------------

\begin{algorithm}[t!]
\caption{\textsc{FindScope}}\label{alg:findscope}
\begin{algorithmic}[1]

\Require $e$, $R$, $Y$, $B$ ($e$: the example, $R$, $Y$: sets of variables, $B$: the bias)
\Ensure $B$, $Scope$ ( $B$: the updated bias, $Scopes$: a set of variables, the scope of a constraint in $C_T$

\Function{FindScope}{$e$, $R$, $Y$, $B$}	

	\If{ $\kappa_B(e_{R}) \neq \emptyset$ } 

		\If{ $ASK(e_R)$ = ``yes'' } $B \leftarrow B  \setminus  \kappa_B(e_R) $;
		\Else \quad \Return $(B, \emptyset)$;
		\EndIf

	\EndIf

	\If{ $|Y| = 1$ } \Return $(B, Y)$;
	\EndIf

	\State split $Y$ into $<Y_1, Y_2>$ such that $|Y_1| = \lceil |Y|/2 \rceil $;

        \If{ $\kappa_B(e_{R \cup Y_1}) = \kappa_B(e_{R \cup Y})$ } $S1 \leftarrow \emptyset$
	\Else \quad $(B, S_1) \leftarrow FindScope(e,R \cup Y_1, Y_2, B)$; 
        \EndIf
        \If{ $\kappa_B(e_{R \cup S_1}) = \kappa_B(e_{R \cup Y})$ } $S2 \leftarrow \emptyset$
        \Else \quad $(B, S_2) \leftarrow FindScope(e,R \cup S_1, Y_1, B)$;
        \EndIf
        
	\State \Return $(B, S_1 \cup S_2)$;

\EndFunction

\end{algorithmic}
\end{algorithm}

% End of algorithm findScope -----------------------------------------

%\textsc{FindScope} achieves a logarithmic complexity in terms of the number of queries posted to the user by splitting the variables approximately in half in each step (line 6).
In practice, the problem that must be solved in each step is to find a set of variables $Y_1 \subset Y$, splitting the previously considered set of variables $Y$ into two parts (line 6). Then, $Y_1$ is used in the next query posted to the user while $Y_2$ is the removed set of variables that can be taken into account in the next queries. Depending on the answer of the user we can then update $B$ and decide which part of the problem to focus on next. Existing \textsc{FindScope} functions naively choose $Y_1 \subset Y$, by splitting the set $Y$ in half. %, keeping into account which sets of variables have previously been used, which are given as parameters in each recursive call. \dt{after all I think the commented part is not needed in our story}
The advantage of this approach is that a logarithmic number of steps is achieved. However, no information about the violated constraints from $B$ is used, and no guidance is utilized. The set of variables to remove, or keep, in the assignment is usually chosen randomly~\cite{bessiere2023learning}. 

The problem of finding a $Y_1 \subset Y$, so that the \textsc{FindScope} procedure is correct and will lead to finding the scope of a violated constraint, can be formulated as:

\begin{equation}
\begin{aligned}
\text{find } Y_1 \text{ s.t. } \emptyset \subsetneq Y_1 \subsetneq Y
\end{aligned}
\end{equation}
That is because, in each query, we get information either for $Y_1$, if the answer of the user remains negative, or for $Y_2$, if the answer of the user changes to positive. Thus we want both $Y_1, Y_2 \supsetneq \emptyset$.%\footnote{Proof of correctness can be found in the appendix.}

This problem can be formulated as a CSP with boolean variables $BV$, with $|BV| = |Y|$, deciding whether a variable $x_i \in Y$ is included in $Y_1$. 
The CSP contains the following constraint: 
\begin{equation}
\label{eq:correct}
\begin{aligned}
0 < \sum_{bv_i \in BV} \llbracket bv_i \rrbracket < |Y|
\end{aligned}
\end{equation}

However, just choosing any (arbitrarily sized) subset of $Y$ can result in many unneeded recursive calls and a large number of queries. Now that we have formally formulated this problem, we can modify the constraints and/or add an objective function in order to improve the performance of \textsc{FindScope}. In order to achieve the logarithmic complexity from \cite{bessiere2023learning}, we can impose the following constraint in our CSP formulation:

\begin{equation}
\label{eq:fs-half}
\begin{aligned}
\sum_{bv_i \in BV} \llbracket bv_i \rrbracket = \left\lfloor \frac{|Y|}{2} \right\rfloor
\end{aligned}
\end{equation}

%\senne{Something confuses me here: we first say: Bessiere's FindScope splits Y in half. Then we say: let's alter the constraint such that $Y_1$ just needs to be a non-empty strict subset of Y. The we say: let's add the constraint that $Y_1$ needs to be half the size of $Y$. Why did we go through this trouble, instead of just saying that still half of the variables need to be chosen, but now an objective function is added to this choice, and then formalizing this as a CSP directly?} \dt{Bessiere's Findscope splits Y in half, not by imposing a constraint and modeling it as a csp, just by saying "take half the variables randomly". So, we first formulate the problem as a csp, and we show what it needs in order to be correct. Then, we show how to exactly do the same thing as in findscope. and then we propose the modification}

We propose to use this CSP formulation of the problem, and to integrate the objective function from~\Cref{eq:guideqgen} to guide \textsc{FindScope} queries. Notice that, the queries asked in \textsc{FindScope} take into account the set $R$, which is the set $R \cup Y_1$ from the previous recursive call, where we split $Y$. In addition, we now are not generating assignments, but deciding which variable assignments from the existing example $e$ to include in the next query $ASK(e_{R \cup Y_1})$, Thus, we propose to maximize the following objective function:

\begin{equation}
\label{eq:guidefs}
\begin{aligned}
\sum_{c \in \kappa_B(e)} \llbracket var(c) \in R \cup Y_1 \rrbracket \cdot (1 - |\Gamma| \cdot \llbracket {\cal M}(c) \rrbracket)
\end{aligned}
\end{equation}

We also slightly modify the constraint from~\Cref{eq:fs-half}, as, when deciding which constraints to violate in the next query, the number of variables these constraints participate in could be lower than half (but still needs to be at least one, as in~\Cref{eq:correct}). As a result, the constraint becomes

\begin{equation}
\begin{aligned}
0 < \sum_{bv_i \in BV} \llbracket bv_i \rrbracket \leq \left\lfloor \frac{|Y|}{2} \right\rfloor
\end{aligned}
\end{equation}

%\senne{I feel like the above is a bit confusing and needs some work. I'll wait with updating until we discuss so that I for sure understand correctly what it's trying to convey.} \dt{I see the confusion with the way it was written. I changed some parts}

\textbf{Correctness}
We now prove that \textsc{FindScope} is still correct when our modification of line 6 is used, as long as the constraint from~\Cref{eq:correct} holds.

\begin{prop}
\label{prop:findscope}

Given the assumption that $C_T$ is representable by $B$,  \textsc{FindScope} (with our modification at line 6) is correct. 
\end{prop}

\begin{proof} 

\textbf{Soundness} We will now prove that given an example $e_Y$, \textsc{FindScope} will return a set of variables $S$, such that there exists at least one violated constraint $c \in C_T$ s.t. $\forall x_i \in S \mid x_i \in var(c)$ .

An invariant of {\em FindScope} is that the example $e$ violates at least one constraint whose scope is a subset of $R \cup Y$ (i.e. ASK($R \cup Y$) = ``no''). 

\begin{equation}
\begin{aligned}
\label{eq:inv1}
\kappa_{C_T}(e_{R \cup Y}) \supsetneq \emptyset
\end{aligned}
\end{equation}

That is because it is called only when the example $e_Y$ is classified as non-solution by the user and the recursive calls at lines 8 and 10 are reached only if the conditions at lines 7 and 9 respectively are false.

In addition, {\em FindScope} reaches line 5 only in the case that $e_R$ does not violate any constraint from $C_T$ (i.e. ASK($e_R$) = ``yes'' at line 3).

\begin{equation}
\begin{aligned}
\label{eq:inv2}
\kappa_{C_T}(e_{R}) = \emptyset
\end{aligned}
\end{equation}

In {\em FindScope} variables are returned (and added in $S$) only at line 5, in the case $Y$ is a singleton. 

\begin{equation}
\begin{aligned}
(\ref{eq:inv1}),(\ref{eq:inv2}) \implies \exists Y^{\prime} \subseteq Y \text{ s.t. } Y^{\prime} \subseteq var(c) \mid c \in \kappa_{C_T}(e) \\ \xRightarrow[\text{line 5}]{|Y| = 1} Y \subseteq var(c) \mid c \in \kappa_{C_T}(e)
\end{aligned}
\end{equation}

Thus, for any $x_i \in S$ we know that $x_i \in var(c) \mid c \in C_T$. 

\textbf{Completeness}
We will now prove that given an example $e_Y$, the set of variables $S$ returned by \textsc{FindScope} will be the full scope of a constraint in $C_T$, i.e. there exists at least one constraint $c \in C_T$ for which $S = var(c)$. 

\textsc{FindScope} in \Cref{alg:findscope} has been proven to be complete in~\cite{bessiere2023learning}. The key part in that is line 6, splitting $Y$ into 2 parts. The requirement is that in no recursive call we end up with $Y = \emptyset$, so that it continues searching in different subsets of variables in each call. This means that in the recursive call of line 9, $Y_2 \neq \emptyset$ and in the recursive call of line 10, we must have $Y_1 \neq \emptyset$.

Due to the constraint imposed in \Cref{eq:correct}, we know that $Y_1 \supsetneq \emptyset$ and also that

\begin{equation}
\begin{aligned}
Y_1 \subsetneq Y \implies Y_1 \subsetneq Y_1 \cup Y_2 \implies Y_2 \neq \emptyset
\end{aligned}
\end{equation}

Thus, this constraint guarantees that $Y_1, Y_2 \neq \emptyset$, meaning that \textsc{FindScope} is still complete.
\end{proof}

\subsection{Guiding FindC}

After the system has located the scope of a violated constraint, it calls function \textsc{FindC}~\cite{bessiere2013constraint,bessiere2023learning} to find the relation of the violated constraint. To locate this constraint, \textsc{FindC} asks partial queries to the user in the scope returned by \textsc{FindScope}. Alternative assignments are used for the variables in the scope given, to discriminate which of the candidate constraints with that scope is part of the target problem. In order to do so, \textsc{FindC} functions currently use the following query generation step:

\begin{equation}
\begin{aligned}
\text{find } e_S^{\prime} \in sol( C_L[S] \wedge \emptyset \subsetneq \kappa_{B}(e^{\prime}_S) \subsetneq \Delta),
\end{aligned}
\end{equation}
with $S$ being the scope found in the previous step and $\Delta$ the set of candidates for this scope, initially being equal to the set of violated constraints in the previous example $\kappa_{B}(e_S)$.

The objective function typically used in this step, in order to again achieve a logarithmic complexity in terms of the number of queries posted, is to try to half the number of violated candidates, minimizing a slack variable $b$ such that 
\begin{equation}
\begin{aligned}
b = \left\lfloor \frac{|\Delta|}{2} \right\rfloor - \kappa_{B}(e^{\prime}_S)
\end{aligned}
\end{equation}

We propose to replace this objective function with one that guides the query generation in the same way as in~\Cref{eq:guideqgen,eq:guidefs}:

\begin{equation}
\label{eq:guidefindc}
\begin{aligned}
\sum_{c \in \Delta} \llbracket e_S \not\in sol(\{c\}) \rrbracket \cdot (1 - |\Gamma| \cdot \llbracket {\cal M}(c) \rrbracket)
\end{aligned}
\end{equation}

%observe the high similarity between the guiding functions (2), (7) and (11) for the three types of query generation in interactive QA
%\section{Using Probabilistic Classification to Guide Constraint Acquisition}

\section{Experimental Evaluation}
\label{sec:exp}

In this section, we perform an experimental evaluation of our proposed approaches, aiming to answer the following research questions:

\begin{itemize}

\item [(Q1)] How well can ML classifiers predict whether a candidate constraint is part of the target constraint network?

\item [(Q2)] What is the effect of using probabilistic classification to guide query generation in CA?

\item [(Q3)] What is the added benefit of also guiding the other layers of CA?

\end{itemize}

\subsection{Benchmarks}
We selected the benchmarks for our experiments to cover different cases, including some puzzle problems that are typically used as benchmarks to evaluate CA systems, some problems closer to real-world applications with a subset of them having a more regular structure, and one randomly generated. The latter was included to evaluate the performance of our system when it cannot learn anything.

More specifically, we used the following benchmarks for the experimental evaluation:

\textbf{Random.}
We used a problem with 100 variables and domains of size 5. We generated a random target network with 495 binary constraints from the language $\Gamma = \{\geq, \leq, <,>,\neq, = \}$. The bias was initialized with 19,800 constraints, using the same language.

\textbf{9x9 Sudoku}
The Sudoku puzzle is a $n^2 \times n^2$ grid ($n=3$ for the case we used), which must be completed in such a way that all the rows, columns, and $n^2$ non-overlapping $n \times n$ squares contain the distinct numbers.  This gives a {\em vocabulary} having 81 variables and domains of size 9. The target constraint network consists of 810 $\neq$ constraints.
The bias was initialized with 12,960 binary constraints, using the language $ \Gamma = \{\geq, \leq, <,>,\neq, = \}$.

\textbf{Jigsaw Sudoku.}
The Jigsaw Sudoku is a variant of Sudoku in which the $3 \times 3$ boxes are replaced by irregular shapes. %We used the instance that is displayed in Figure \ref{fig:jigsaw}, 
It consists of 81 variables with domains of size 9. The target network contains 811 binary $\neq$ constraints on rows, columns, and shapes. 
The bias $B$ was constructed using the language $\Gamma = \{\geq, \leq, <,>,\neq, = \}$ and contains 19,440 binary constraints.

\hide{
\textbf{Job-shop scheduling.}
The job-shop scheduling problem involves scheduling a number of jobs, consisting of several tasks, across a number of machines, over a certain time horizon. %Each task can only be run on a specific machine. 
The decision variables are the start and end times of each task.
%Each task can only be run on a specific machine, so the task-machine assignment is implied and not reflected in the decision variables or the constraints.
There is a total order over each job's tasks, expressed by binary precedence constraints. %These constraints express that a task can only start once all of the tasks that must precede it have ended. 
There are also constraints capturing the duration of the tasks and that tasks should not overlap in the same machine.
%Each task has also has a duration. There are duration constraints that capture that a task's end time must equal its start time plus its duration. Finally, there are non-overlap constraints that capture that two tasks that are run on the same machine cannot overlap in time. 
The language $\Gamma = \{\geq, \leq,$ \mbox{$<,$} $>,\neq, =, x_i + c = x_k \}$ was used, with $c$ being a constant from 0 up to the max. duration of the jobs. We used a problem instance containing 10 jobs, 3 machines (i.e. $|X| = 60$), and a time horizon of 15 steps, leading to a bias containing 14,160 constraints.
%\senne{The job-shop scheduling problem is now introduced with much more detail than the other problems. Perhaps we can drop the description that starts with `Each task can only ...'.}
}

\textbf{Exam Timetabling}
There are $ns$ semesters, each containing $cps$ courses, and we want to schedule the exams of the courses in a period of $d$ days, On each day we have $t$ timeslots and $r$ rooms available for exams. The variables are the courses ($|X| = ns \cdot cps$), having as domains the timeslots they can be assigned on ($D_i = 1, ..., r \cdot t \cdot d$).
There are $\neq$ constraints between each pair of exams. % because they cannot be assigned in the same timeslot of a day in the same room. 
Also, two courses in the same semester cannot be examined on the same day, %. As the slots per day are $spd = t*r$, this can be 
which is expressed by the constraints 
$\lfloor x_i/spd \rfloor \neq \lfloor x_j/spd \rfloor$, $\forall i,j$ in the same program. 
We used an instance with $ns = 8$, $cps = 6$, $d = 10$ and $r = 3$.
%We used an instance with 8 semesters, 6 courses per semester, with 10 days of exams, 3 timeslots per day and 3 rooms available. 
This resulted in a model with 48 variables and domains of size 90. $C_T$ consists of $1,128$ constraints. The language given is $\Gamma = \{\geq, \leq, <,>,\neq, =, \lfloor x_i/spd \rfloor \neq \lfloor x_j/spd \rfloor\}$, creating a bias of size 7,896.

\textbf{Nurse rostering}
There are $n$ nurses, $s$ shifts per day, $ns$ nurses per shift, and $d$ days. The goal is to create a schedule, assigning a nurse to all existing shifts. The variables are the shifts, and there are a total of $d \cdot s \cdot ns$ shifts. The variables are modeled in a 3D matrix. The domains of the variables are the nurses ($D^{x_i} = \{1, ..., n\} $). Each shift in a day must be assigned to a different nurse and the last shift of a day must be assigned to a different nurse than the first shift of the next day. In the instance used in the experiments, we have $d = 7$, $s = 3$ and $ns = 5$ . The available nurses are $n = 18$.  This results in $|X| = 105$ with domains $\{1, ..., 18\}$. $C_T$ consists of 885 $\neq$ constraints. The bias was built using the language $\Gamma = \{\geq, \leq, <,>,\neq, = \}$, resulting in $|B| = 32,760$.

\subsection{Experimental settings}

%\begin{itemize}

%\item
All the experiments were conducted on a system carrying an Intel(R) Core(TM) i7-2600 CPU, 3.40GHz clock speed, with 16 GB of RAM. 
%\item 
The guiding techniques are integrated within \textsc{GrowAcq}, utilizing \textsc{MQuAcq-2} as the underlying algorithm. We compare our approach with the counting method introduced in~\cite{tsouros2023guided}, referred to as ``count``, as well as with \textsc{GrowAcq} without guiding (``base``). In the latter, the objective in query generation is simply to maximize the number of violated candidate constraints.
%\item 
%\item 
We use the following classifiers: Random Forests (RF), Gaussian Naive Bayes (GNB), Multi-layer Perceptron (MLP), and Support Vector Machines (SVM).\footnote{Hyperparameter tuning details can be found in the appendix.} 

All methods and benchmarks were implemented\footnote{Our code is available online at: https://github.com/Dimosts/ActiveConLearn} in Python using the CPMpy constraint programming and modeling library~\cite{guns2019increasing}. OR-Tools' CP-SAT~\cite{ortools} solver was used.
For query generation, we used PQ-Gen from [Tsouros, Berden, Guns, 2023], with a cutoff of 1 second to return the best query found. Implementation of the ML classifiers was carried out using the Scikit-Learn library~\cite{scikitlearn}.

The comparison is based on the following two metrics: 
\begin{itemize}
    \item \textit{the number of queries} (\# of Queries), which is very important for the applicability of interactive CA systems in real-world scenarios,
    \item \textit{the maximum user waiting time} (Max T), which is of paramount importance when human users are involved.
\end{itemize}

The results presented in each benchmark, for each algorithm, are the means of 10 runs.

%\end{itemize}

\subsection{Results}
% Q1 ML classifiers predict constraints (also increasing percentages of dataset)
\subsubsection{Q1}

\begin{figure}[tb]
\centering
     \includegraphics[width=0.4\textwidth]{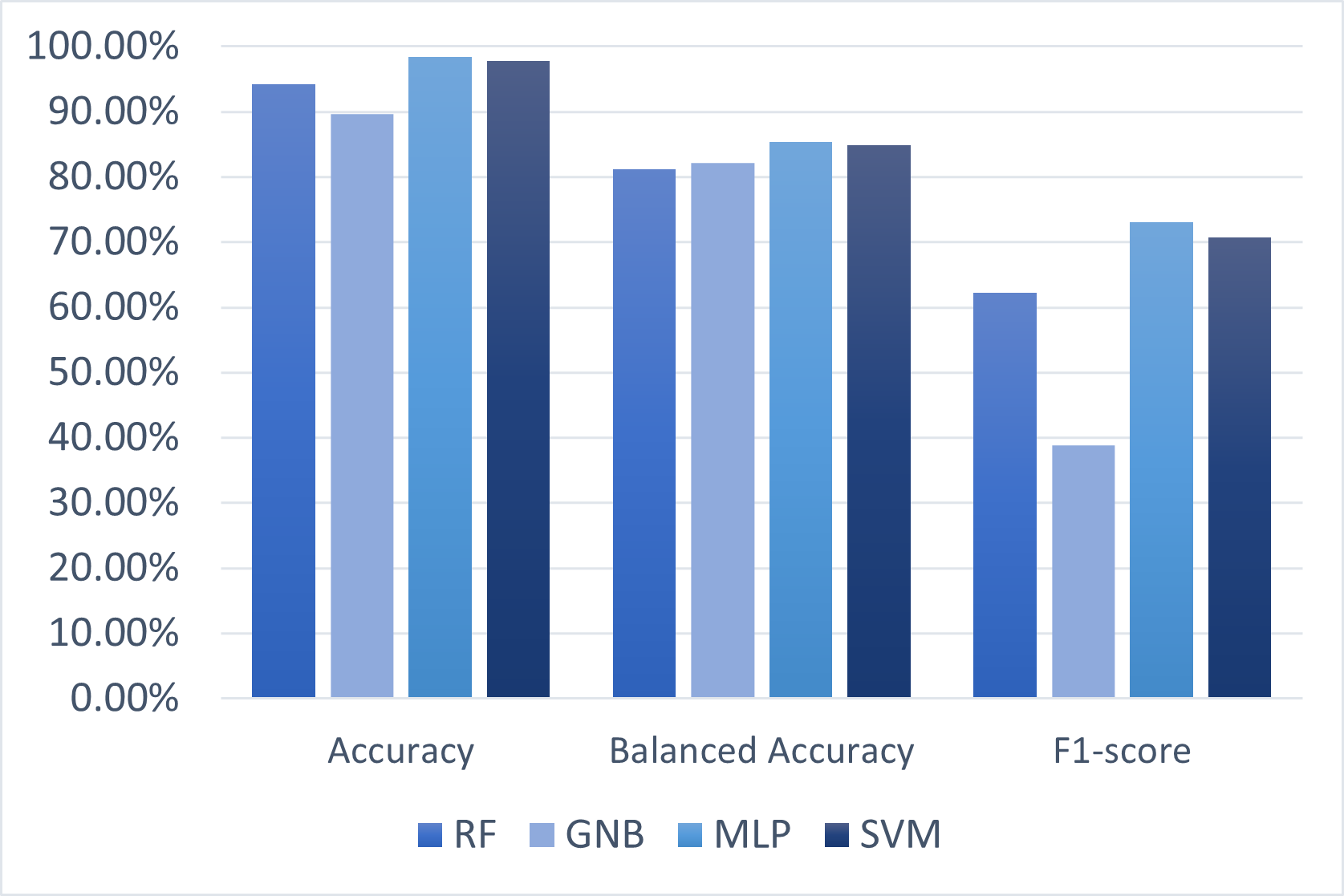}	
     \caption{Classification results with different classifiers}
 \label{fig:classification}
\end{figure}

In order to answer this question, we performed a 10-fold cross-validation with each classifier on all the datasets, and present the averages. As metrics for the comparison, we use \textit{Accuracy}, \textit{Balanced Accuracy} (the datasets are highly unbalanced, with $<10\%$ of $B$ typically having a positive label), and \textit{F1-score}. The results are shown in~\Cref{fig:classification}.

\begin{figure*}[ht]
\centering
     \begin{subfigure}[b]{0.3\textwidth}
        \captionsetup{justification=centering}
         \centering
         \includegraphics[width=\textwidth]{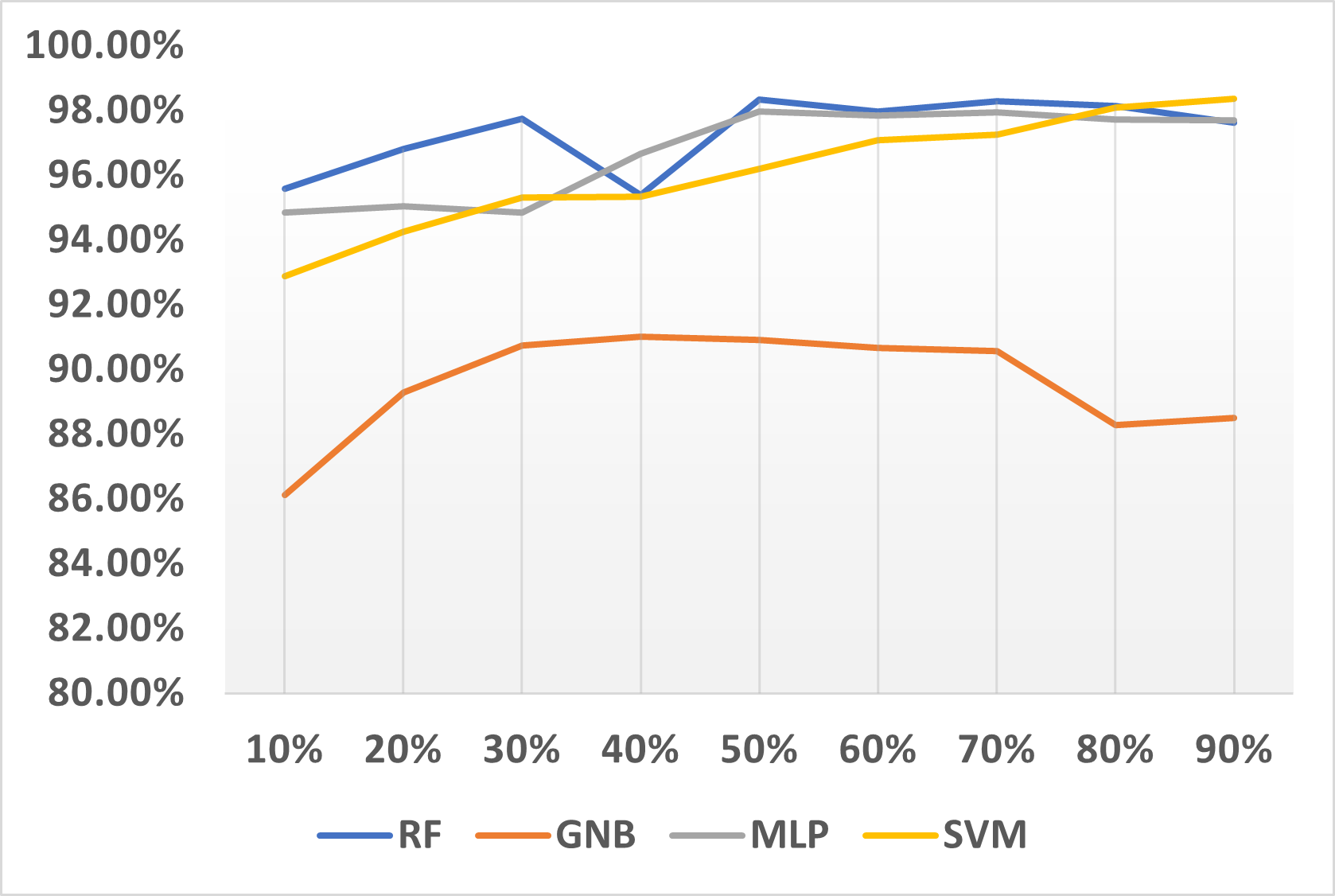}
         \caption{Accuracy}
     \end{subfigure}
    \begin{subfigure}[b]{0.3\textwidth}
        \captionsetup{justification=centering}
         \centering
         \includegraphics[width=\textwidth]{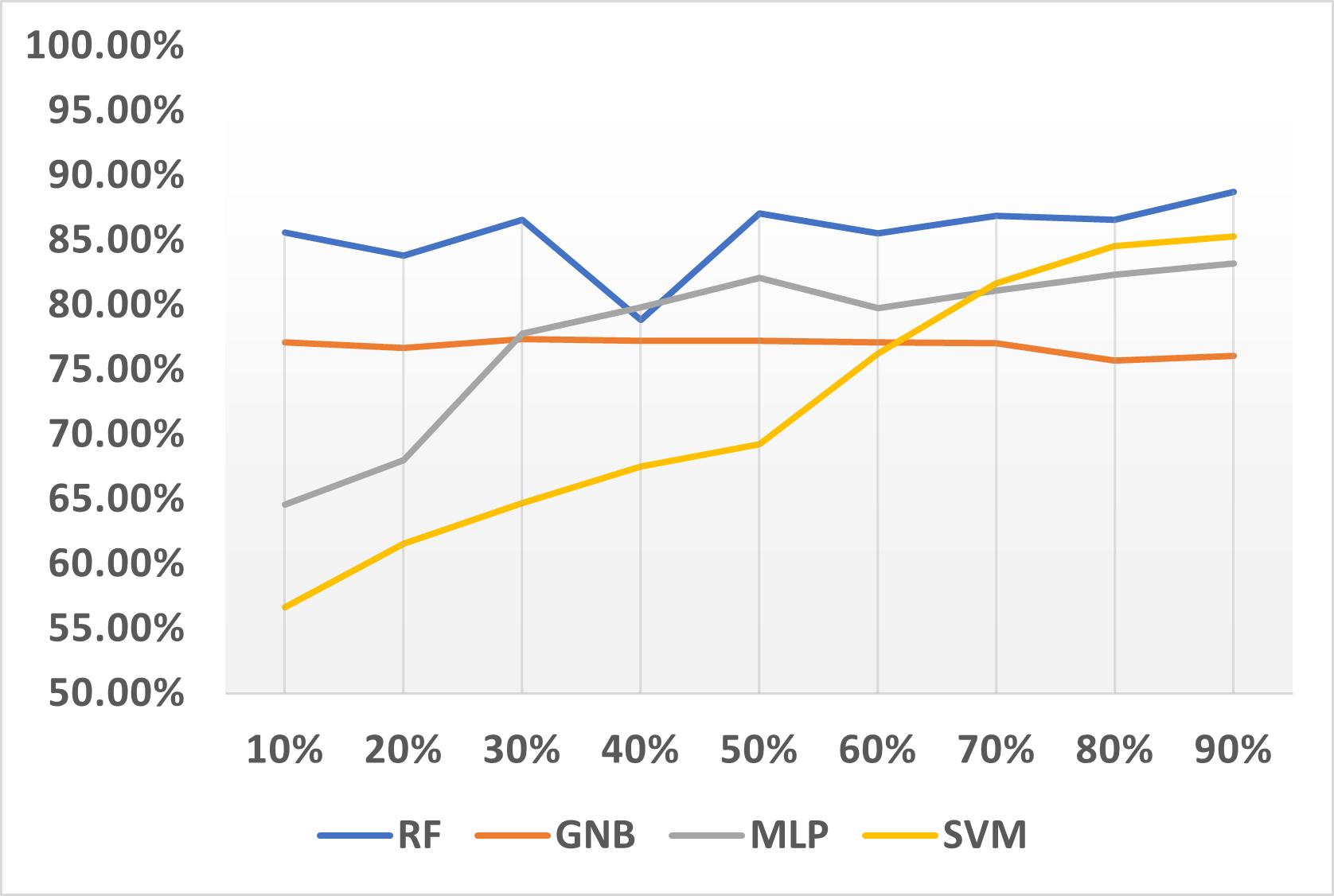}
         \caption{Balanced Accuracy}
     \end{subfigure}
    \begin{subfigure}[b]{0.3\textwidth}
        \captionsetup{justification=centering}
         \centering
         \includegraphics[width=\textwidth]{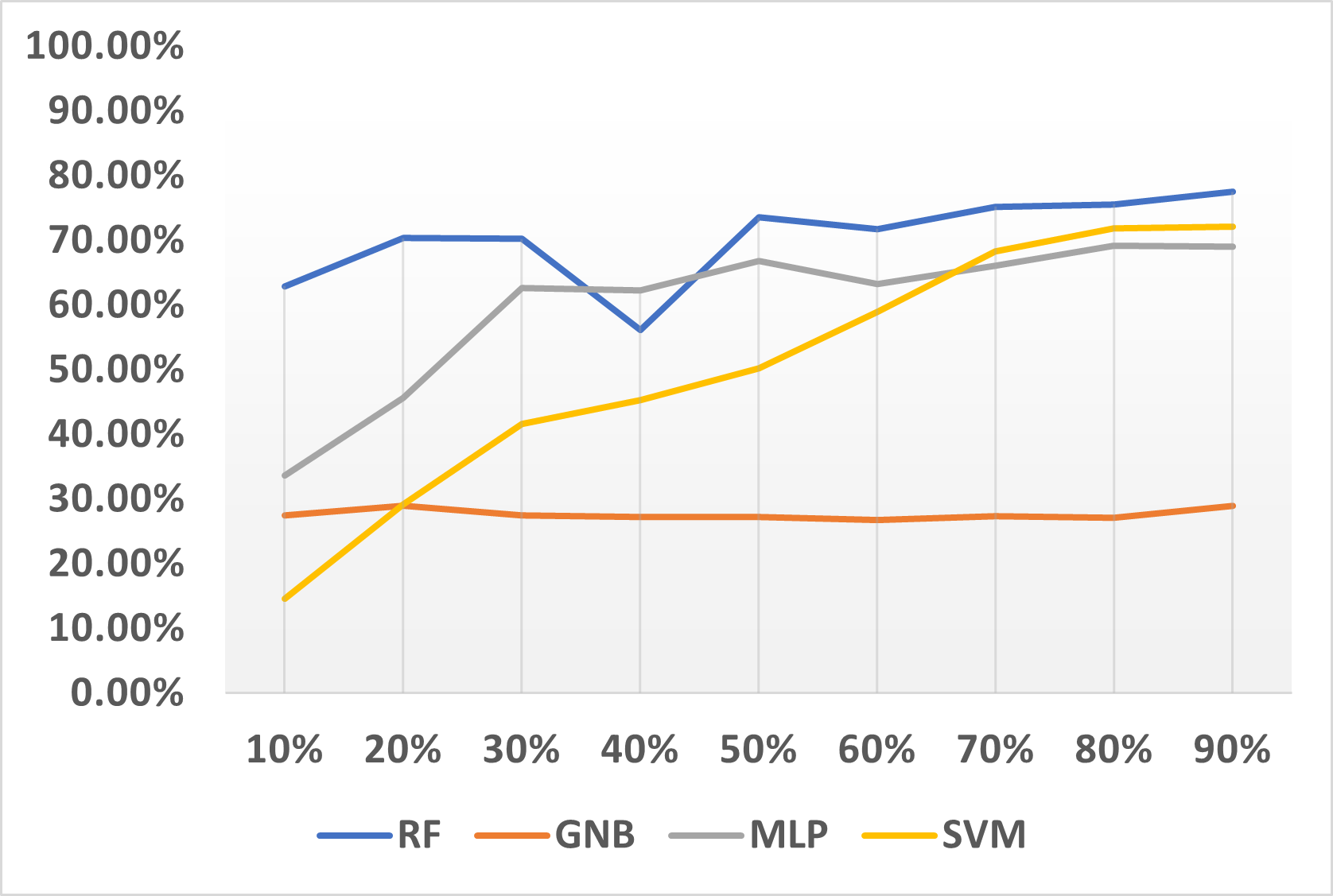}
         \caption{F1 Score}
     \end{subfigure}
	
	\caption{Classification results when only part of the dataset is available for training}
 \label{fig:percentage_classification}
\end{figure*}

We can notice that all classifiers considered achieve a decent accuracy and balanced accuracy, with GNB performing slightly worse than the rest, and MLPs performing best. Focusing on F1-score, GNB presents quite bad results, but the rest of the classifiers still achieve a score higher than 70\%. The results indicate that based on the way the dataset of constraints is created and the features used, it is possible to successfully train and use ML models to predict whether a constraint is part of the target network or not.

However, in order to use the classifiers to assist during the acquisition process -- guiding it to generate promising queries, based on the predictions -- it is of high importance to evaluate how they perform not only when the labels for all candidate constraints are available, but when only some parts of the dataset are available (as this is the case during the CA process). Thus, we conducted an experiment to evaluate how the classifiers perform when only a percentage of the dataset is available. We used an increasing portion of the dataset as the training set, to evaluate their performance in different stages of the acquisition process, with the rest of the candidates being the test set. The order of the constraints in the dataset was decided based on the order in which they were added in 5 different runs of CA systems. The averages are presented in~\Cref{fig:percentage_classification}.

We can observe that RF achieves the best results in all metrics in the beginning when only a small portion of the dataset is labeled, with MLP and SVMs reaching the same performance only when most of the dataset is available. GNB is shown here to have a bad performance throughout the process, having very low accuracy and F1-score.

% Q2 Effect of using (proba) classification to guide qgen
\subsubsection{Q2}

\begin{figure}[tb]
\centering
     \begin{subfigure}[b]{0.4\textwidth}
        \captionsetup{justification=centering}
         \centering
         \includegraphics[width=\textwidth]{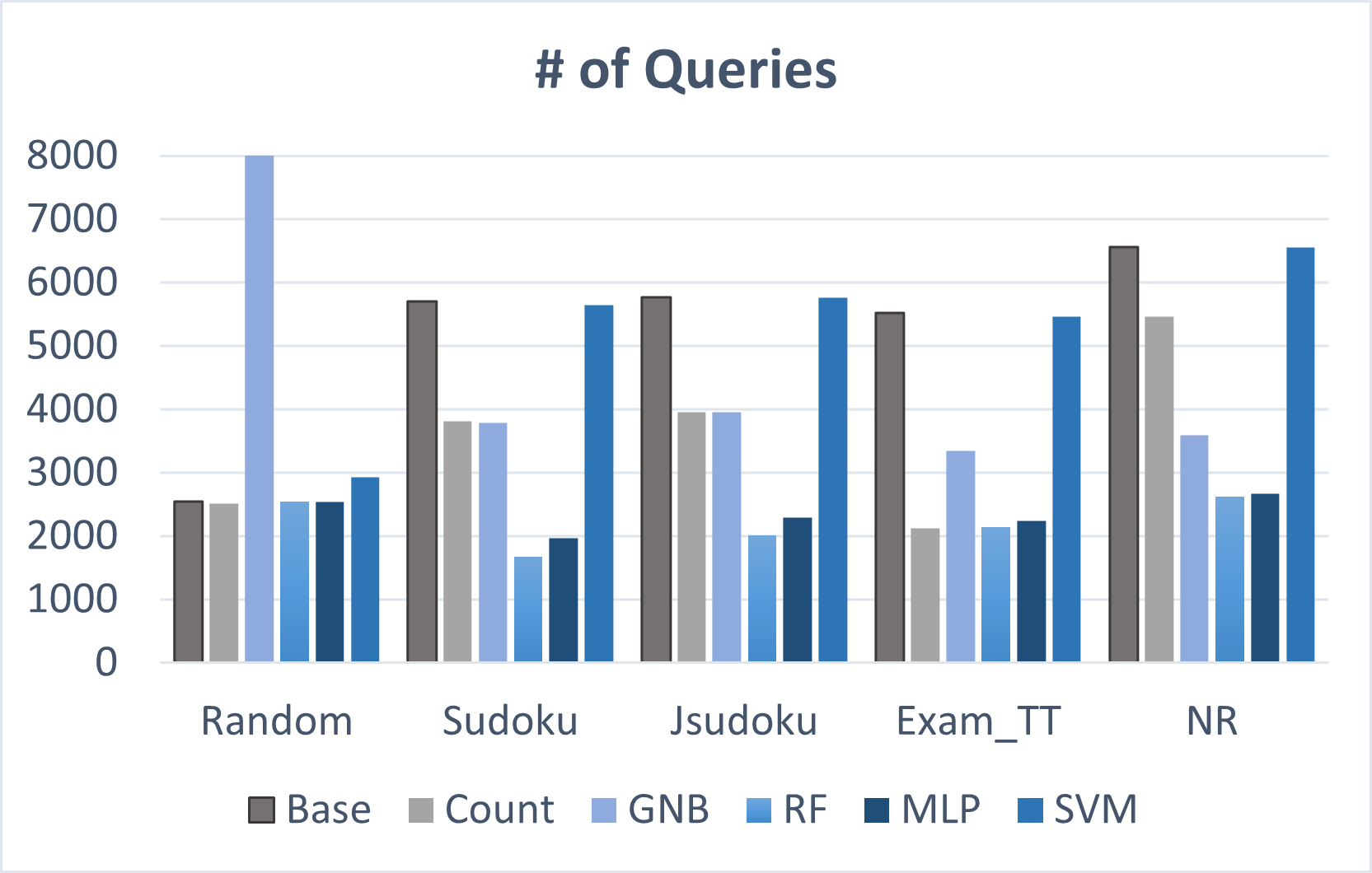}
         \caption{}
     \end{subfigure}
    \begin{subfigure}[b]{0.4\textwidth}
        \captionsetup{justification=centering}
         \centering
         \includegraphics[width=\textwidth]{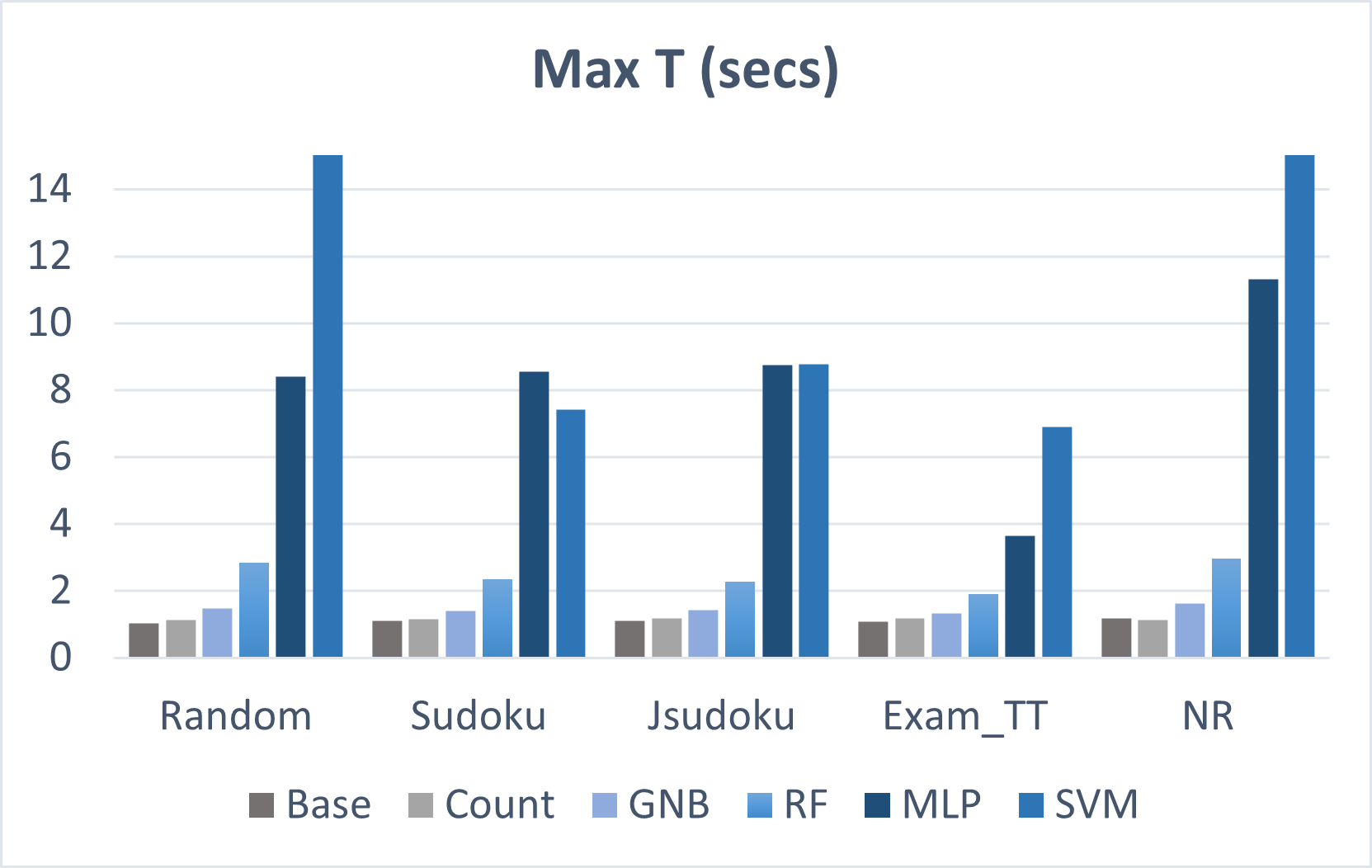}
         \caption{}
     \end{subfigure}
	
	\caption{Results when guiding query generation using probabilistic classification}
 \label{fig:proba_oracle}
\end{figure}

Let us now focus on the effect of using probabilistic classification to guide query generation in CA. \Cref{fig:proba_oracle} presents the result when using the different classifiers, compared to guiding using the simple counting method from~\cite{tsouros2023guided} (Count) and the GrowAcq without guiding (Base).

In all benchmarks, except Random and JSS, the decrease in the number of queries is significant compared to both the baseline and the simple counting method, for most classifiers. When SVMs are used, the performance is similar to the baseline because %many configurations for SVMs that presented better results were rejected in the tuning process due to their bad training time performance. The selected configuration 
it has a lower accuracy in earlier stages of the acquisition process and thus does not offer any meaningful guidance early enough. GNB presents decent results in some benchmarks, but its overfitting is shown in the Random benchmark, where guiding should not detect any patterns and thus have a similar performance as the baseline, which is true for the rest of the classifiers.
Using RF and MLPs is the most promising, giving the best results in all benchmarks, with RF being superior in some cases. We attribute RF's superior performance to the fact that it already achieves good prediction performance when only a small portion of the constraints is labeled, i.e., at the beginning of the acquisition process (\Cref{fig:percentage_classification}).

Regarding the waiting time for the user, it includes 1 second for query generation (based on the imposed cutoff), and the rest of the waiting time involves mainly the training and prediction time. As a result, we can see higher waiting times when SVM or MLPs are used, which need a larger training time, while GrowAcq with no guiding (Base) and the simple counting method (Count) have similar waiting times because they do not need any training. We can also observe that the training time for GNB and RF is small and very reasonable for interactive settings, as the maximum observed waiting time is less than 2s.

Overall, we can see that using RF to predict probabilities for the candidate constraints, and then guide query generation based on these predictions, seems to be the best choice, both in terms of the number of queries and the user waiting time. It can decrease the number of queries required by up to 70\% compared to the baseline (and up to 56\% compared to the counting method), with the average decrease in the benchmarks that have structure (i.e., all except Random) being 52\% (and 32\% compared to counting). At the same time, the increase in the user waiting time is minor and acceptable for interactive scenarios.

% Q3 Guiding all layers together!
\subsubsection{Q3}

\begin{figure}[tb]
\centering
     \begin{subfigure}[b]{0.4\textwidth}
        \captionsetup{justification=centering}
         \centering
         \includegraphics[width=\textwidth]{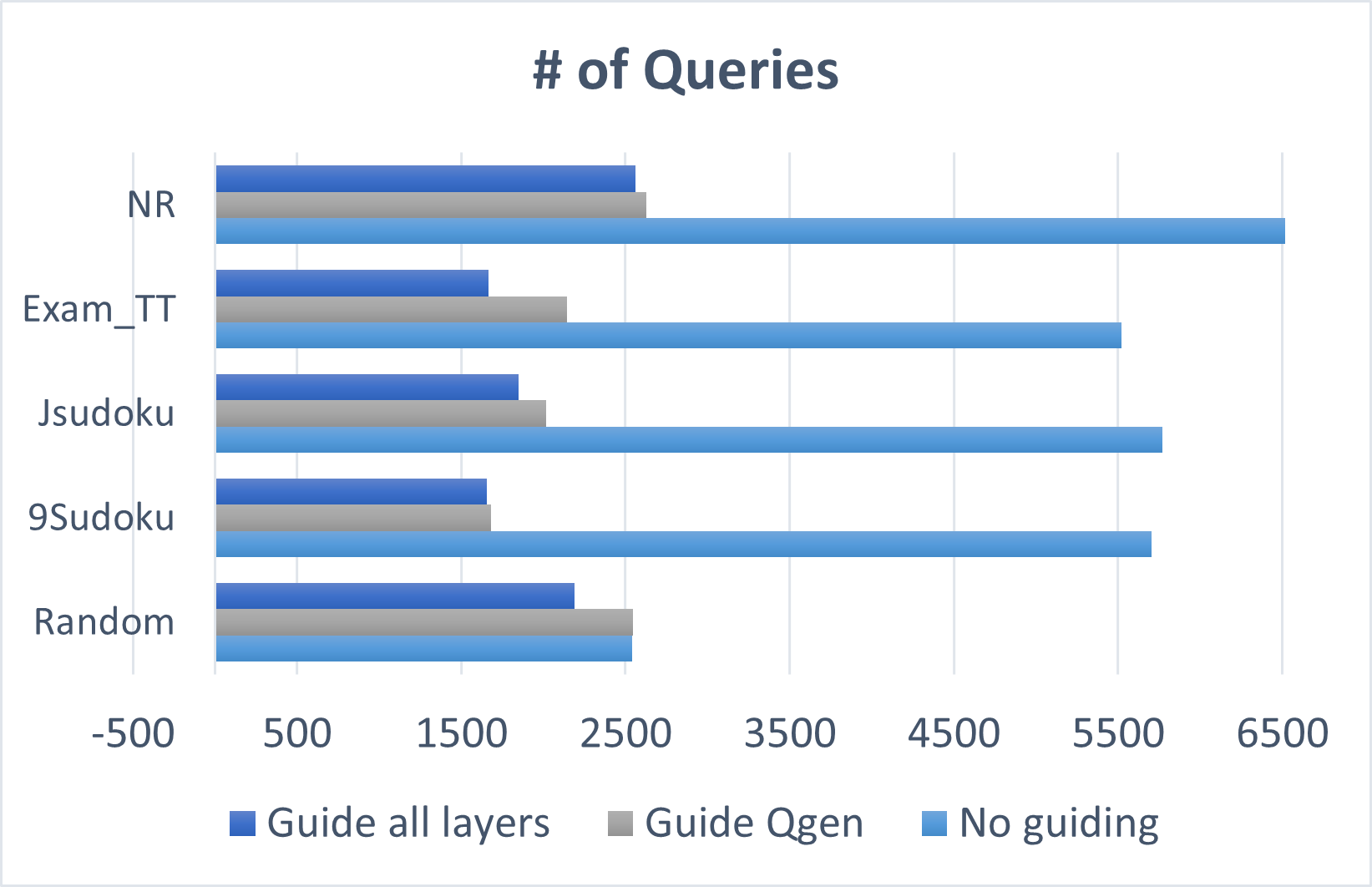}
         \caption{}
     \end{subfigure}
    \begin{subfigure}[b]{0.4\textwidth}
        \captionsetup{justification=centering}
         \centering
         \includegraphics[width=\textwidth]{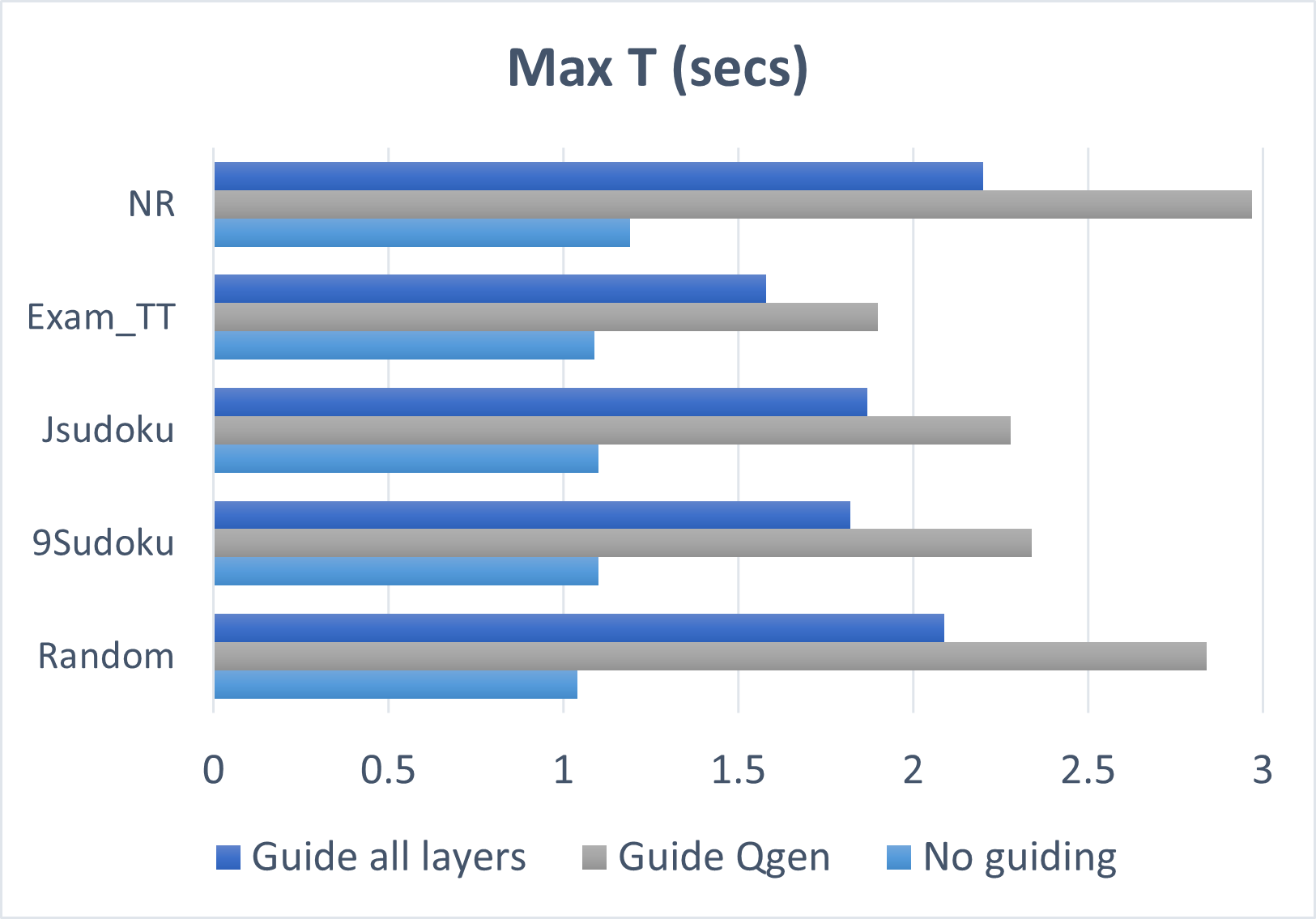}
         \caption{}
     \end{subfigure}
	
	\caption{Evaluating the effect of guiding all layers of CA}
 \label{fig:gfs}
\end{figure}

We now evaluate the effect of also guiding the other layers of interactive CA where queries are asked to the user. We only use RF, as it presented the best performance in the previous experiment, and we compare it against the baseline (without guiding) and against only guiding query generation. \Cref{fig:gfs} presents the results. 

We can see a comparatively small but additional improvement when guiding all layers compared to guiding only top-level query generation. The improvement is relatively small because guiding query generation already led to significantly fewer queries needed in \textsc{FindScope} and \textsc{FindC}. However, the additional decrease still goes up to 22\% (in Exam TT), with the average decrease in the number of queries reaching 10\%. In addition, we can observe a slight reduction in the maximum waiting time for the user when we use the predictions to guide all layers of CA. We believe that this is because by prioritizing the removal of candidate constraints in all layers, $B$ is shrinking during \textsc{FindScope} and \textsc{FindC}. Thus, fewer top-level queries will be needed leading to a smaller amount of retraining steps.

%\senne{This is a good explanation, but one I feel that we should evidence. One way of doing so is to also include results for guiding \textit{only} \textsc{FindScope} and \textsc{FindC}, without guiding query generation} \dt{Not sure we want to go towards that direction, cause then we may have to even start presenting all combinations of guiding/not guiding. Why then not present guiding with only findscope and then only findc? The purpose of this question, is to show what is the additional effect of guiding the rest of the layers, when we guide query generation (and also show the overall effect compared to no guiding). Not to show the separate effect of each}

\section{Conclusions}
\label{sec:concl}

One bottleneck of major importance in interactive CA is the number of queries needed to converge. The search-based learning used in CA is often not able to detect patterns existing in the problem while learning the constraints, and thus, does not use such patterns to better guide its search. In this work, we tighten the connection of ML and CA, by using for the first time statistical ML methods that can learn during the acquisition process and predict whether a candidate constraint is part of the problem or not. We propose to use probabilistic classification, using the predictions from the ML models in order to guide the search process of CA. 
In doing so, we extend recent work that guided query generation using probabilities derived via a simple counting-based method.
We also extend guidance to the other components of CA that post queries to the user, further reducing the number of queries. Our experimental evaluation showed that the number of queries was decreased by up to 72\% compared, greatly outperforming the state-of-the-art. These findings confirm that using statistical ML methods can indeed detect patterns in constraint models, while they are being learned. This can be a stepping stone to further reducing the number of queries in interactive CA.

Future work should investigate the use of online learning in this setting, as data becomes available gradually. Other opportunities include learning a prior distribution over constraints and transfer learning across different problems. We also think that our closer integration with statistical ML techniques can be a stepping stone towards being able to handle wrong answers from the user, which is an important part of future work, in order to make interactive CA more realistic. 
Finally, extending interactive CA systems to be able to learn global constraints and linear inequalities with constants is very important for the efficacy and efficiency of such systems.

\begin{quote}
\begin{small}
\bibliography{paper}
\end{small}
\end{quote}

%\section{Acknowledgments}
%...

\newpage
\appendix
\section{Tuning details}
We used RF and GNB in their default settings, while we tuned the most important hyperparameters for MLP and SVM.
For tuning, we used the final dataset for all benchmarks, having labeled all candidate constraints. A grid search, coupled with 10-fold cross-validation, was conducted, using balanced accuracy as the metric to address class imbalance. Hyperparameter combinations surpassing a 10-second training time were omitted to ensure relevance in interactive scenarios.

Regarding SVMs, we explored kernel types (linear, polynomial, and RBF) and regularization parameter (C) values [0.01, 0.1, 1, 10, 100, 1000, 10000]. For the RBS kernel we also explored different values for gamma (['auto', 'scale']). The optimal configuration was identified as an SVM with an RBF kernel, C=100, and gamma='scale'. 

For the MLPs, we focused on tuning the number of hidden layers (1-2), the number of neurons per layer ([8, 16, 32, 64]), and the learning rate ([0.001, 0.01, 0.1, 1]). The optimal configuration was identified as an MLP with a single hidden layer having 64 neurons and with a learning rate of 0.1, using ReLu as the activation function.

\end{document}